\documentclass{article}

    \PassOptionsToPackage{numbers, compress}{natbib}

    \usepackage[preprint]{neurips_2024}

\usepackage[utf8]{inputenc} 
\usepackage[T1]{fontenc}    
\usepackage{hyperref}       
\usepackage{url}            
\usepackage{booktabs}       
\usepackage{amsfonts}       
\usepackage{nicefrac}       
\usepackage{microtype}      
\usepackage{xcolor}         
\usepackage{colortbl}
\usepackage{amsmath,amssymb,amsfonts}
\usepackage{graphicx}
\usepackage{subfigure}
\usepackage{multirow}
\usepackage{listings}
\usepackage{pythonhighlight}
\usepackage{paralist}
\usepackage{arydshln}
\usepackage{pifont}
\usepackage{wrapfig}
\usepackage{caption}
\usepackage{natbib}
\usepackage{algorithm}
\usepackage{algorithmic}
\usepackage{wrapfig}
\usepackage{amsmath}
\usepackage{subcaption}
\usepackage{multirow}
\usepackage{booktabs}

\definecolor{lightblue}{RGB}{50,150,220}
\hypersetup{
    colorlinks=true,
    linkcolor=red,
    citecolor=cyan,
    filecolor=magenta,      
    urlcolor=magenta,
    }

\def\method{BeamVQ}
\title{\method{}: Aligning Space-Time Forecasting Model via Self-training on Physics-aware Metrics}

\author{%
  Hao Wu \\
  Tencent, TEG \\
  \texttt{easyluwu@tencent.com} \\
  \And
  Xingjian Shi \\
  Boson AI \\
  \texttt{xshiab@connect.ust.hk}\\
  \And
  Ziyue Huang \\
  Tencent, TEG \\
  \texttt{ziyuehuang@tencent.com} \\
  \And
  Penghao Zhao \\
  Tencent, TEG \\
  \texttt{hymiezhao@tencent.com} \\
  \And
  Wei Xiong \\
  Tsinghua University \\
  \texttt{xiongw21@mails.tsinghua.edu.cn} \\
  \And
  Jinbao Xue \\
  Tencent, TEG \\
  \texttt{jinbaoxue@tencent.com} \\
  \And
  Yangyu Tao \\
  Tencent, TEG \\
  \texttt{brucetao@tencent.com} \\
  \And
  Xiaomeng Huang \\
  Tsinghua University \\
  \texttt{hxm@tsinghua.edu.cn} \\
  \And
  Weiyan Wang \\
  Tencent, TEG \\
  \texttt{neowywang@tencent.com} \thanks{Corresponding author.}
}

\begin{document}

\maketitle

\begin{abstract}
Data-driven deep learning has emerged as the new paradigm to model complex physical space-time systems.
These data-driven methods learn patterns by optimizing statistical metrics and tend to overlook the adherence to physical laws, unlike traditional model-driven numerical methods. Thus, they often generate predictions that are not physically realistic. On the other hand, by sampling a large amount of high quality predictions from a data-driven model, some predictions will be more physically plausible than the others and closer to what will happen in the future. Based on this observation, we propose \emph{Beam search by Vector Quantization} (\method{}) to enhance the physical alignment of data-driven space-time forecasting models. The key of \method{} is to train model on self-generated samples filtered with physics-aware metrics. To flexibly support different backbone architectures,
\method{} leverages a code bank to transform any encoder-decoder model to the continuous state space into discrete codes.
Afterwards, it iteratively employs beam search to sample high-quality sequences, retains those with the highest physics-aware scores, and trains model on the new dataset. 
Comprehensive experiments show that \method{} not only gave an average statistical skill score boost of more than 32\% for ten backbones on five datasets, but also significantly enhances physics-aware metrics.
\end{abstract}

\section{Introduction}
Accurate modeling of complex physical systems is essential for various scientific and engineering domains such as meteorology, climatology, and environmental science. Traditionally, these systems are solved using numerical methods~\cite{jouvet2009numerical, rogallo1984numerical, orszag1974numerical, griebel1998numerical}, which employ discrete approximation techniques to solve sets of equations derived from physical laws. Although these physics-driven methods ensure compliance with fundamental principles such as conservation laws~\cite{karpatne2017theory,karnopp2012system,pukrushpan2004control}, they require highly trained professionals for development~\cite{lam2022graphcast}, incur high computational costs~\cite{pathak2022fourcastnet}, are less effective when the underlying physics is not fully known~\cite{takamoto2022pdebench}, and cannot easily improve as more observational data become available~\cite{lam2022graphcast}.

Recently, data-driven deep learning starts to revolutionize the space of space-time forecasting for complex physical systems~\cite{gao2022earthformer, wu2024earthfarsser, li2020fourier, tan2022simvp, shi2015convolutional, pathak2022fourcastnet, wu2023pastnet,bi2023accurate,lam2022graphcast,zhang2023skilful}. Rather than relying on differential equations governed by physical laws, the data-driven approach constructs a model by optimizing statistical metrics such as Mean Squared Error (MSE), using large-scale datasets. These methods~\cite{wang2022predrnn, shi2015convolutional, wang2018eidetic, tan2022simvp, gao2022earthformer, wu2024earthfarsser} are orders of magnitude faster, and excel in capturing the intricate patterns and distributions present in high-dimensional nonlinear systems~\cite{pathak2022fourcastnet}. Despite their success, purely data-driven methods fall short in generating physically plausible predictions, leading to unreliable outputs that violate critical physical theorem~\cite{bi2023accurate, pathak2022fourcastnet, wu2024neural}.

Previous works have tried to combine physics-driven methods and data-driven methods to get the best of both worlds. Some methods try to embed physical constraints in the neural network~\cite{long2018pde,greydanus2019hamiltonian,cranmer2020lagrangian,guen2020disentangling}. For example, PhyDNet~\cite{guen2020disentangling} adds a physics-inspired PhyCell in the recurrent network. However, such methods require explicit formulation of the physical rules along with specialized designs for network architectures or training algorithms. As a result, they lack flexibility and cannot easily adapt to different backbone architectures. Another type of methods~\cite{raissi2019physics,li2021physics,hansen2023learning}, best exemplified by the Physics-Informed Neural Network (PINN)~\cite{raissi2019physics}, leverages physical equations as additional regularizers in neural network training~\cite{hansen2023learning}. Physics-Informed Neural Operator (PINO)~\cite{li2021physics} extends the data-driven Fourier Neural Operator (FNO) to be physics-informed by adding soft regularizers in the loss function. However, PDE-based regularizers impose optimization challenges~\cite{krishnapriyan2021characterizing,wang20222} that often lead to poor solutions. More recently, PreDiff~\cite{gao2024prediff} trains a latent diffusion model for probabilistic forecasting, and guides the model's sampling process with a physics-informed energy function. However, PreDiff requires training a separate knowledge alignment network to integrate the physical constraints, which is not needed in our method.

\begin{figure*}[t]
    \vspace{-15pt}
    \centering
    \includegraphics[width=\textwidth]{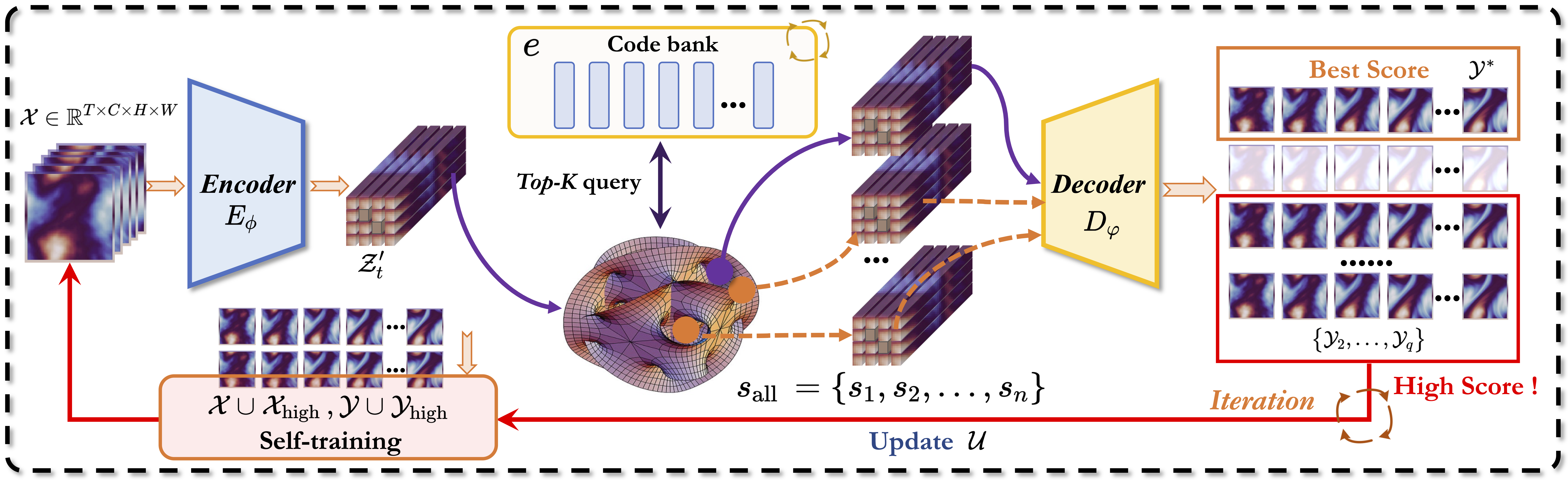}
    \caption{
    \textbf{Overview of \method{}:} Input data $\mathcal{X}$ is fed to the encoder $E_{\phi}$ to produce the latent state $\mathcal{Z}_t = E_{\phi}(\mathcal{X})$. \method{} inserts code bank quantization layers between the encoder $E_{\phi}$ and decoder $D_{\varphi}$. $K$ candidate latent states are sampled via top-$K$ beam search: $s_{all} = \{s_1, s_2, \ldots, s_{n}\}$. Each state $s_i$ is decoded to the output sequence: $\mathcal{Y}_i = D_{\varphi}(s_{i}), \forall i \in [N]$. These candidate outputs are filtered based on the physics-aware score $F(\mathcal{Y}_i)$, in which the high-score samples are added back to the training dataset. \method{} iteratively generates new samples with the updated model weights to shift the data distribution to better match the physical law.
    }
    \label{fig:Idea_main} 
    \vspace{-20pt}
\end{figure*}

In this paper, we propose \emph{Beam search by Vector Quantization} (\method{})~\cite{steinbiss1994improvements, van2017neural,razavi2019generating}, a novel framework designed to enhance the physical consistency of space-time forecasting models. \method{} is based on the following observation: if we draw a large amount of high quality samples from a probabilistic forecasting model, some samples will be more physically consistent than the others and closer to the real future. Thus, \method{} iteratively trains the model on self-generated samples filtered by physics-aware metrics. To ensure our method is flexible to different backbones, we leverage a code bank to discretize the continuous state space, transforming any encoder-decoder-based forecasting model into probabilistic model. To ensure the quality of the generated samples, we extend the nearest neighbor code book lookup to top-K lookup, thereby facilitating beam search sampling. The overall workflow of \method{} is illustrated in Figure~\ref{fig:Idea_main}. The design of \method{} makes it applicable to a variety of backbone architectures and physical constraints, and does not have optimization challenges as soft-constraint-based methods. Our self-training method is also relevant to the recent progress in aligning large-scale generative models with human preferences through methods like reward-reranked finetuning~\cite{dong2023raft} and iterative preference optimization~\cite{yuan2024self}. However, our goal diverges significantly as we aim for \emph{physical alignment}, i.e., refining space-time forecasting model to produce predictions that are physically real, setting it apart from methods focused on aligning with human preferences.

Our experiments demonstrate that \method{} consistently boosts performance across 10 different space-time forecasting models, achieving state-of-the-art results on 5 benchmarks. Moreover \method{} improves both statistical metrics and physics-aware metrics, and is adaptable to long-term forecasting scenarios. This makes \method{} a powerful tool to integrate physical knowledge into data-driven deep learning methods, opening new avenues for research and application in fields such as aerospace, biomedical engineering, and meteorology.

\section{Preliminaries}
We first discuss our problem setting and then introduce the background of physics-aware metric.

\subsection{Space-Time Forecasting} 
Our objective is to predict the future of a space-time system with historical data. We represent the input as a four-dimensional tensor $\mathcal{X} \in \mathbb{R}^{T \times C \times H \times W}$, where $T$ denotes the number of time steps, $C$ is the number of physical observations, and $H$ and $W$ are spatial dimensions. The output $\mathcal{Y}$ is a tensor $\mathcal{Y} \in \mathbb{R}^{T' \times C' \times H' \times W'}$ that represents the future observation. We employ a probabilistic model $\mathcal{P}(\mathcal{Y} | \mathcal{X}; \theta)$, where $\theta$ represents the model parameters, to generate possible future sequences $\{\mathcal{Y}_1, \mathcal{Y}_2, \ldots, \mathcal{Y}_{T'}\}$. Here, $\mathcal{Y}_i \in \mathbb{R}^{C' \times H' \times W'}$ depicts the output at the $i$-th prediction step.

\subsection{Physics-aware Metrics}
\label{sec:physics-aware metrics}

\begin{figure*}[h]
    \centering
    \includegraphics[width=\textwidth]{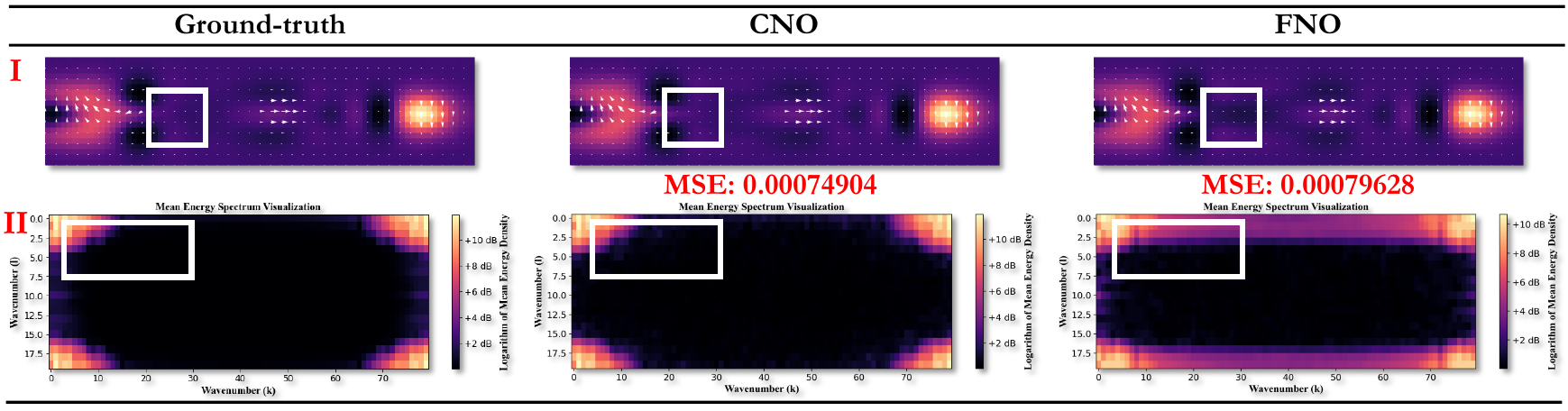}
    \caption{\textcolor{red}{\textbf{I}}. The top row shows the actual and predicted distributions of height and speed with a white box highlighting significant discrepancies. Dotted patterns reflect the direction and magnitude of velocity vectors. \textcolor{red}{\textbf{II}}. The second row shows the average energy spectrum, depicting energy distribution across wavenumbers (k) and highlighting fluctuations in the dynamic system over different scales. Although CNO and FNO achieve similar MSE, the energy spectrum of CNO is closer to the ground-truth and is thus more physically realistic.}
    \label{fig:metric} 
        \vspace{-10pt}
\end{figure*}

Although MSE has been widely used as the loss function in training and evaluation, it can hardly capture the physical inconsistency. Figure~\ref{fig:metric} gives an example. In the figure, both CNO~\cite{raonic2024convolutional} and FNO~\cite{li2020fourier} have low MSE and are perceptually similar with the Ground-truth ($7.49\times 10^{-4}$ and $7.96\times 10^{-4}$ respectively), but CNO has much more realistic energy specturm than FNO. This emphasizes the importance of \textit{\textbf{evaluating models based not only on statistical metrics but also on physical consistency}}. Statistical metrics like MSE overlook a model's compliance with physical laws, leading to unreliable predictions even when the metric is low. Therefore, it is crucial to establish physics-aware metrics alongside statistical metrics to ensure more robust model performance.

In this paper, we consider three different physics-aware metrics for dynamic system applications. These metrics collectively describe turbulence characteristics, assess fluid behavior and kinetic energy distribution in the physical field. Following are the details of these physics-aware metrics:

\textbf{Divergence of the Velocity Field}~\cite{tuckerman1989divergence}: In the context of fluid dynamics, particularly in the study of turbulence, an incompressible flow implies that the divergence of the velocity field must be zero. This constraint ensures mass conservation within the flow. Mathematically, the divergence of a velocity field $\mathbf{w}$ at a point $\mathbf{x}$ is given by $\nabla \cdot \mathbf{w}(\mathbf{x})$. For a discrete representation, the average divergence across the entire field can be calculated as follows:
\begin{equation}
    \text{Divergence} = \frac{1}{M} \sum_{j=1}^M\left|\nabla \cdot \mathbf{w}_j\right|,
\end{equation}
where $ M $ is the total number of points in the prediction step, and $ \mathbf{w}_j $ is the velocity vector at point $ j $.

\textbf{Turbulence Kinetic Energy (TKE)}~\cite{nagata2013turbulence}: In fluid dynamics, turbulence kinetic energy represents the average kinetic energy per unit mass of eddies in turbulent flow. Physically, it is defined by the root mean square of measured velocity fluctuations:
\begin{equation}
\left(\overline{\left(u^{\prime}\right)^{2}}+\overline{\left(v^{\prime}\right)^{2}}\right) / 2, \quad \overline{\left(u^{\prime}\right)^{2}}=\frac{1}{T} \sum_{t=0}^{T}(u(t)-\bar{u})^{2},
\end{equation}
and $t$ is the time step. We calculate the turbulence kinetic energy for predicted velocity fields.

\textbf{Energy Spectrum}~\cite{gutzwiller1970energy}: The turbulence energy spectrum, denoted as $E(k)$, describes how the kinetic energy of turbulence is distributed among various scales of motion, corresponding to different sizes of vortices or eddies. It is defined by the following integral:
\begin{equation}
\int_{0}^{\infty} E(k) \, dk = \frac{1}{2} \left( \overline{(u')^2} + \overline{(v')^2} \right),
\end{equation}
where $ k $ represents the wavenumber, indicating spatial frequency in the Fourier transform domain. This integral assessment helps to understand the energy ratio between large and small eddies, which is crucial for characterizing the turbulence's behavior.

\section{Methodology}
\subsection{Framework Overview}
Generally, \method{} iteratively applies three steps: beam search, updating the dataset by adding physically-plausible samples, and self-training. 
The detailed process is described as follows: \textbf{(1) Beam search}: At each searching step, the probabilistic model generates a set of candidate states and assigns likelihood scores. The search space is pruned by only keeping the top-k states with high physics-aware scores at each step.  
\textbf{(2) Dataset update:} To maintain physical consistency, \method{} employs physics-aware metrics (e.g., velocity field divergence, turbulence kinetic energy, and energy spectrum) to evaluate all output sequences from beam search. 
Output sequences with top physics-aware scores are added to the training dataset. This step shifts the training data distribution to better match physical laws. 
\textbf{(3) Self-training:} \method{} updates the model with the latest training dataset for better alignment with physics-aware scores.

\begin{figure*}[t]
\vspace{-15pt}
    \centering
    \includegraphics[width=\textwidth]{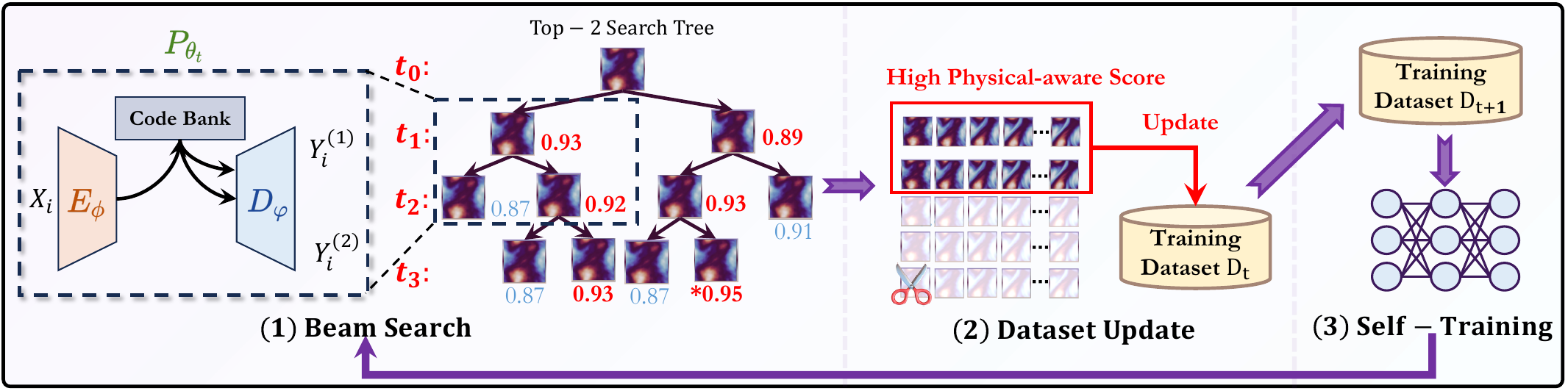}
    \vspace{-15pt}
    \caption{The closed-loop of \method{}: \textbf{(1) Beam search.} it generates different output sequences for each input, by sampling different continuous states with the probabilistic model. 
  \textbf{(2) Dataset Update:} it filters out the output sequences with high physics-aware Scores to expand the training dataset.
  \textbf{(3) Self-training:} it trains on the shifted data distribution for better physical consistency.}
    \label{fig:Beam} 
    \vspace{-10pt}
\end{figure*}

This iterative process continuously optimizes the model's predictive capabilities as well as the alignment with physics-aware metrics. 
Figure~\ref{fig:Beam} illustrates how all three steps are interconnected,
forming a closed-loop optimization process that gradually enhances the space-time forecasting performance.


In every step of beam search, the input sequence $\mathcal{X}$ passes through the encoder $E_{\phi}$ to generate latent states. 
Through sampling different entries in the code bank, the probabilistic model $P_{\theta_t}$ generates candidate states $s_i$ to replace the original one. The decoder $D_{\varphi}$ decodes these candidates into predictions, which are selected based on the physics-aware metric $F$ for the next level of the beam search tree. In the final level, Beam search $B$ selects the highest-scored sequence $\mathcal{Y}^*$ as the final output, namely $\mathcal{Y}^* = B\left(\{F(D_{\varphi}(s_i)) \mid s_i \in P_{\theta_t}(E_{\phi}(\mathcal{X}))\}\right)$. 
And \method{} collects all output sequences higher than a threshold to iteratively expand the training dataset $D_{t}$. Formally speaking, we have $D_{t+1} = D_t \cup D_{\text{high}}$.
Then \method{} continues to update the model as $\theta_{t+1} \leftarrow \text{$\mathcal{U}$}(\theta_t, \text{$\mathcal{L}$}(D_{t+1}))$, where $D_{t+1}$ is the updated dataset and $L$ denotes the statistical loss function. 
In the following subsections, we introduce all the details about \method.

\subsection{Probabilistic Model and Beam Search}

Inspired by the success of beam search in discrete token sequence generation~\cite{steinbiss1994improvements,boulanger2013high,guo2023beam} (e.g., natural language, musical notes, and chemical formula), we propose a general method to convert the existing deterministic model into a probabilistic model to sample different frame states in continuous space.

\textbf{Encoder:} The encoder $\mathcal{Z}_t = E_{\phi}(\mathcal{X})$ computes latent vectors from raw observation data in historical inputs.
As a general method,
\method{} can employ any popular backbone networks as the encoder, such as ViT~\cite{dosovitskiy2021an}, Earthfarseer~\cite{wu2024earthfarsser}, SimVP~\cite{tan2022simvp}, FNO~\cite{li2020fourier}, and CNO~\cite{raonic2024convolutional}. 
The encoder takes high-dimensional data of physical systems $\mathcal{X}_t \in \mathbb{R}^{C \times H \times W}$ and maps it to the latent vector space $\mathcal{Z}_t \in \mathbb{R}^{l \times D}$ through a series of transformations, where $l$ is the number of tokens and $d$ is the dimension of each token. Next, $\mathcal{Z}_t$ is mapped to a new vector space $\mathcal{Z}'_t \in \mathbb{R}^{l \times d}$ by dimension reduction to match the dimension of code bank.
This transformation process can be described as:
\begin{equation}
\mathcal{Z}_t = E_{\phi}(\mathcal{X}_t) \quad, \quad {\mathcal{Z}}_t^{\prime}=\sigma\left(W \cdot \mathcal{Z}_t+b\right),
\end{equation}
where $\mathcal{Z}_t^{\prime}$ is the projection of $\mathcal{Z}_t$ in the low dimension, $W$ and $b$ are the projection matrix and bias, and $\sigma$ is the activation function.

\textbf{Beam Search Sampling \& Decoder:} 
Unlike previous works that make deterministic predictions, we follow VQVAE to plug in a code bank to build a probabilistic model. Furthermore, we enhance it by substituting nearest-neighbor search with Top-K calculations for beam search. This improvement allows the model to generate multiple potential states and build a beam search tree whose node stands for a candidate state(i.e., a time frame) as shown in Figure~\ref{fig:Beam}(1). 

\begin{wrapfigure}{r}{7cm}
\vspace{-8pt}
 \centering
 \includegraphics[width=0.4\textwidth]{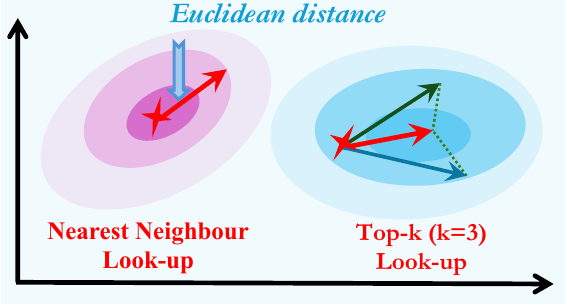}
 \vspace{-5pt}
 \caption{Visualisation of the embedding space.}
 \label{fig:vqvae}
\end{wrapfigure}

As shown in Figure~\ref{fig:vqvae}, traditional VQVAE models select the encoding vector with the smallest Euclidean distance to the input vector in the quantization step. However, \method{} modifies this approach by choosing the Top-K closest encoding vectors. 
Following the simplified notation in VQVAE~\cite{van2017neural,razavi2019generating}, we have Equation~\ref{eq:eq5} to show how to select the Top-K vectors $ \{e_{k_1}, e_{k_2}, \ldots, e_{k_K}\}$ close to the current state $\mathcal{Z}'_t$ for the candidate states in the next tree level. 
\begin{equation}
\label{eq:eq5}
    \{\mathcal{Z}_{t+1}^{k_1}, \ldots, \mathcal{Z}_{t+1}^{k_K}\} = \{e_{k_1}, \ldots, e_{k_K}\}, \quad \text{where } \{k_1,\ldots, k_K\} = \text{top-K}_j \|\mathcal{Z}'_t - e_j\|^2,
\end{equation}
where $\mathcal{Z}_{t+1}^{k_i}$ is the latent representation of the $k_i$-th candidate state,  $\text{top-K}_j \|\mathcal{Z}'_t - e_j\|^2$ identifies the indices $\{k_1, k_2, \ldots, k_K\}$ of top K nearest vectors.

For the top-K beam search, we always keep the active node number as K in all levels of the beam search tree(except the initial one). 
Given the K state nodes in any level in the tree, we sample different K candidate states in the next level for each node.
To output the next frame for all $K^2$ candidate sequences,
we feed these states into the decoder $D_{\varphi}$ that can be any popular one (e.g., the deconvolution network).
We evaluate the physics-aware scores of all candidate sequences and only leave the top-K high ones to be active in the future search. 
For any time step $t$, \method{} consider all $K$ active frames $\mathcal{Y}_{t}^{k_p}$ to generate and filter out the new $K$ active candidate frames $\mathcal{Y}_{t+1}^{k_q}$ as:
\begin{equation}
    \mathcal{Y}_{t+1}^{k_q} = \text{BeamVQ}(\mathcal{Y}_{t}^{k_p}), \quad \text{for } p,q \in \{1, \ldots, K \}.
\end{equation}
\method{} repeatedly apply the same process to keep $K$ active candidate sequences until it reaches the maximum tree depth $T'$. 
We consider all paths from the initial level and the final level in the searching tree to generate predicted physical field sequence candidates $\{\mathcal{Y}^1, \mathcal{Y}^2, \ldots, \mathcal{Y}^\mathcal{K}\}$.
Finally, we select the candidate sequence  $\mathcal{Y}^*$ with the highest physics-aware score from the beam search as the inference output.

\textbf{Tradeoff for Long-term Forecasting:} 
Since beam search brings in extra costs,
we can predict multiple time frames together in one step as the tradeoff for long-term forecasting.
If the beam size is $K$, the original strategy samples $ K $ states for each candidate sequence in every step, and it only continues to extend the top-K high-quality sequences in the next step. The total number of computed states is $ S_{\text{original}} = n K^2 $. Assuming that $ K = 5 $ and $ n = 100 $ in long-term forecasting, the number of computed states is about $ 2500$. The optimized strategy combines frames (e.g., 10 frames) and predicts them in parallel, reducing the total number of computed states to be $250$. Namely, it saves about 90\% computations and greatly improves the efficiency of long-term prediction. Our experiments in Section~\ref{sec:long} show it still achieves high prediction quality despite the tradeoff for efficiency.

\subsection{Iterative Self-Training}
\label{subsec:training}
To better align with physics-aware scores, we employ an iterative self-training strategy to train \method{}. Although it still uses the original statistic loss during the training, \method{} filers out its output sequences with high physics-aware scores to iteratively shift data distribution to better match Physical law.

\textbf{Training Loss:} To train the whole model, we minimize the MSE error on the whole training dataset. Additionally, to optimize the code bank using standard gradient descent, we incorporate the stop gradient operator \textbf{sg()}. This operator works a marker during network forward propagation and blocks gradient calculation during backpropagation. Thus, we can express the loss objective as follows:
\begin{equation}
\mathcal{L} = \lambda \left(\frac{1}{TI} \sum_{t=1}^{T} \sum_{i=1}^{I} (\mathcal{Y}^*_{t,i} - \mathcal{Y}_{t,i})^2\right) + \beta \| \mathcal{Z}'_t -  \mathbf{sg}[e]  \|_2^2 + \gamma \| \mathbf{sg}[\mathcal{Z}'_t] - e \|_2^2.
\end{equation}
The loss function includes three weight parameters \(\lambda, \beta, \gamma\) to adjust the contribution of different components to the total loss.

\textbf{Training Details:} At the beginning of training ($t < \mathcal{E}$), the model is trained using only the original training dataset $\mathcal{D}$, with the update formula:
\begin{equation}\label{eq:ori}
    \theta^{(t+1)} = \theta^{(t)} - \eta \nabla_\theta \mathcal{L}(\mathcal{X}, \mathcal{Y}; \theta^{(t)}),
\end{equation}
where $\theta^{(t)}$ are the $t$th cycle parameters, $\mathcal{L}$ is the loss function, and $\eta$ is the learning rate.

Starting from the $\mathcal{E}$th epoch,
\method{} first conducts inference on the all input sequences  $\{\mathcal{X}^1, \mathcal{X}^2, \ldots\ , \mathcal{X}^{N} \}$  to generate their top-K output candidates with beam search. 
Then we evaluate all outputs with physics-aware metrics and add the highest ones and the ones above the threshold to the self-training dataset $\mathcal{D}_{\text{high}}$:
\begin{equation}
\label{eq:q}
\mathcal{D}_{\text{high}}^{(t+1)} = \{\mathcal{X},\mathcal{Y}^i \mid \mathcal{Y}^i= \mathcal{Y}^* \text{ or } F(\mathcal{Y}^i) \geq \text{threshold}, \mathcal{Y}^i \in \mathcal{P}(\mathcal{Y} \mid \mathcal{X}; \theta^{(t)})\} \cup \mathcal{D}_{\text{high}}^{(t)} .
\end{equation}
This process ensures that the model can continuously learn from high-quality data that conforms to the laws of physics while gradually increasing the frequency of self-training, thus improving its predictions' accuracy and physical consistency. It is worth emphasizing that once $\mathcal{X}_{\text{high}}$ and $\mathcal{Y}_{\text{high}}$ are selected, they are always included in the training set for subsequent model self-training.

Then in every self-training epoch,
self-training data $\mathcal{D}^{t+1}_{\text{high}}=( \mathcal{X}^{t+1}_{\text{high}},  \mathcal{Y}^{t+1}_{\text{high}})$ is introduced, 
which contains iteratively added high-quality predictive data. 
The model update formula is adjusted to:
\begin{equation}\label{eq:self}
    \theta^{(t+1)} = \theta^{(t)} - \eta ( \nabla_\theta \mathcal{L}(\mathcal{X} \cup \mathcal{X}^{t+1}_{\text{high}}, \mathcal{Y} \cup \mathcal{Y}^{t+1}_{\text{high}}; \theta^{(t)}).
\end{equation}

\section{Experiment}
\label{sec:experiment}
In this section, we conduct experiments to assess the effectiveness of our method. We cover 5 benchmarks and 10 backbone models. The experiments aim to investigate the following research questions: \textit{{\textbf{RQ1.}}} Can~\method{} enhance the performance of the baselines? \textit{\textbf{RQ2.}} Can~\method{} have better physical alignment? \textit{\textbf{RQ3.}} Can~\method{} produce long-term forecasting? 

\subsection{Experimental Settings}
\begin{figure*}[t]
    \centering
    \begin{subfigure}
        \centering
        \captionof{table}{Performance comparison of various models with and without the BeamVQ method across five benchmark tests (SWE(u), RBC, NSE, Prometheus, SEVIR), using MSE as the evaluation metric. We bold-case the entries with lower MSE. ``Improvement'' represents the average percentage improvement in MSE achieved with BeamVQ.}
        \vspace{-5pt}
        \tiny
        \label{tab:mainres}
        \vskip 0.1in
        \centering
        \begin{scriptsize}
            \begin{sc}
                \renewcommand{\multirowsetup}{\centering}
                \setlength{\tabcolsep}{2.5pt} 
                \begin{tabular}{l|cc|cc|cc|cc|cc}
                    \toprule
                    \multirow{4}{*}{Model} & \multicolumn{10}{c}{Benchmarks}  \\
                    \cmidrule(lr){2-11}
                    & \multicolumn{2}{c}{SWE (u)} & \multicolumn{2}{c}{RBC} & \multicolumn{2}{c}{NSE} &  \multicolumn{2}{c}{Prometheus} &  \multicolumn{2}{c}{SEVIR}   \\
                    \cmidrule(lr){2-11}
                   & Ori & + BeamVQ & Ori & + BeamVQ & Ori & + BeamVQ & Ori & + BeamVQ & Ori & + BeamVQ \\
                    \midrule
                    ResNet &0.0076 & \textbf{0.0033} & \textbf{0.1599} &0.2561 & 0.2330 &\textbf{0.1663}  & 0.2356 &\textbf{0.1987} &  0.0671&  \textbf{0.0542} \\
                    ConvLSTM &0.0024 &\textbf{0.0016}  &0.2726  & \textbf{0.0868} & 0.4094& \textbf{0.1277} & 0.0732 &\textbf{0.0533}  & 0.1757 & \textbf{0.1283}  \\
                    Earthformer & 0.0135&\textbf{0.0093}  & 0.1273 &\textbf{0.1093} &1.8720  &  \textbf{0.1202} &0.2765  &\textbf{0.2001}  &0.0982  & \textbf{0.0521}  \\
                    SimVP-v2 & 0.0013&\textbf{0.0010}  & 0.1234 & \textbf{0.1087} & 0.1238 &\textbf{0.1022}  & 0.1238 &\textbf{0.0921}  &0.0063&  \textbf{0.0032}\\
                    TAU &0.0046 & \textbf{0.0031} & 0.1221 &\textbf{0.0965}  & 0.1205 &\textbf{0.1017} & 0.1201 &\textbf{0.0899}& 0.0059 &  \textbf{0.0029} \\
                    Earthfarseer &0.0075 &\textbf{0.0059}  & 0.1454 &\textbf{0.1023}  & 0.1138 &\textbf{0.0987}  & 0.1176 &\textbf{0.1092}&  0.0065  &  \textbf{0.0021}  \\
                    FNO & 0.0031&  \textbf{0.0024}& 0.1235 & \textbf{0.1053} & 0.2237 & \textbf{0.1005} & 0.3472 & \textbf{0.2275} & 0.0783 &  \textbf{0.0436} \\
                    NMO &0.0021 &\textbf{0.0004}  &0.1123  & \textbf{0.1092} & 0.1007 & \textbf{0.0886} &0.0982  &\textbf{0.0475}  & 0.0045 & \textbf{0.0029}  \\
                    CNO & 0.0146&  \textbf{0.0016}& 0.1327 & \textbf{0.1086} &0.2188 & \textbf{0.1483} &0.1097  &  \textbf{0.0254}&0.0056 &  \textbf{0.0053}  \\
                    FourcastNet &0.0065 &\textbf{0.0061}  & 0.0671 & \textbf{0.0219} & 0.1794 &\textbf{0.1424}  &0.0987  &\textbf{0.0542}  & 0.0721 &\textbf{0.0652}   \\
                    \midrule
                     Improvement(\%)& \multicolumn{2}{c|}{+39.08$\%$}  &  \multicolumn{2}{c|}{+18.97$\%$}   &   \multicolumn{2}{c|}{+35.83$\%$}   &  \multicolumn{2}{c|}{+33.65$\%$}   &  \multicolumn{2}{c}{+35.27$\%$}   \\
                    \bottomrule
                \end{tabular}
            \end{sc}
        \end{scriptsize}
    \end{subfigure}
    \hfill
    \begin{subfigure}
        \centering
        \includegraphics[width=\textwidth]{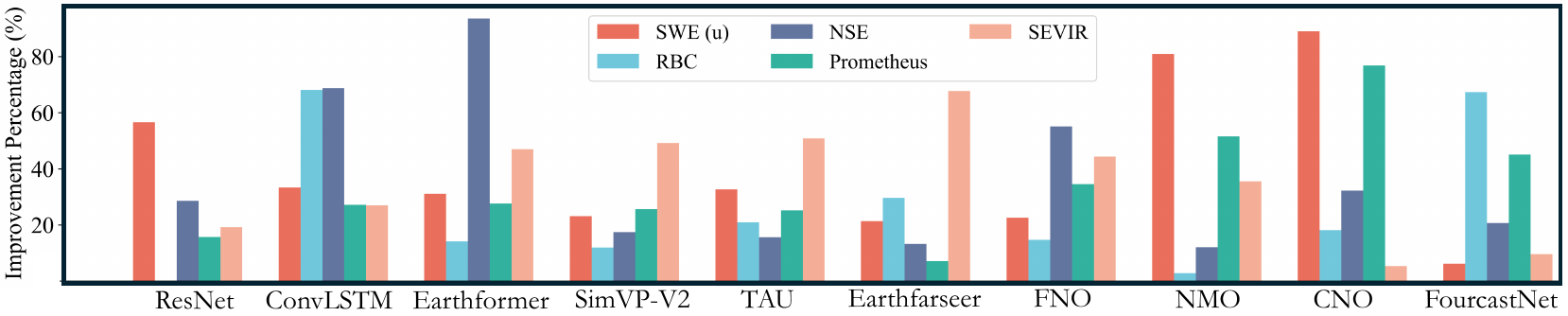}
        \vspace{-10pt}
        \caption{This bar chart shows the improvement percentage of different backbones across multiple benchmarks (SWE (u), RBC, NSE, Prometheus, SEVIR). Each color represents a benchmark. The horizontal axis lists the models, and the vertical axis shows the improvement percentages.}
        \label{fig:improve} 
    \end{subfigure}
    \vspace{-15pt}
\end{figure*}

\textbf{Benchmarks \& Backbones.} Our dataset spans multiple spatiotemporal dynamical systems, summarized as follows: \textbf{$\bullet$ Real-world Datasets}, including SEVIR~\cite{veillette2020sevir}; \textbf{$\bullet$ Equation-driven Datasets}, focusing on PDE~\cite{takamoto2022pdebench} (Navier-Stokes equations, Shallow-Water Equations) and Rayleigh-Bénard convection flow~\cite{wang2020towards}; \textbf{(3) Computational Fluid Dynamics Simulation Datasets}, namely Prometheus~\cite{wu2024prometheus}.  We select core models from three different fields for analysis. Specifically: \textbf{$\bullet$ Spatio-temporal Predictive Learning}, we choose ResNet~\cite{he2016deep}, ConvLSTM~\cite{shi2015convolutional}, Earthformer~\cite{gao2022earthformer}, SimVP-v2~\cite{tan2022simvp}, TAU~\cite{tan2023temporal}, and Earthfarseer~\cite{wu2024earthfarsser} as representative models; \textbf{$\bullet$ Neural Operator}, we compare models like FNO~\cite{li2020fourier}, NMO~\cite{wu2024neural} and CNO~\cite{raonic2024convolutional}; \textbf{$\bullet$ Large Scientific Model}, we study FourcastNet~\cite{pathak2022fourcastnet}. 

\textbf{Implementation details.} Our method trains with MSE loss, uses the ADAM optimizer~\cite{kingma2014adam}, and sets the learning rate to $10^{-3}$. We set the batch size to 100. The training process early stops within 500 epochs. 
Additionally, we set our code bank size as $1024\times 64$, beam size $K$ as 5 or 10, and the threshold as the first quartile of all candidate's scores, which we find suitable for all backbones. 
We implement all experiments in PyTorch~\cite{paszke2019pytorch} on 16 NVIDIA A100-PCIE-40GB GPUs.

\subsection{Assessing The Efficacy Of \method{} (\textit{RQ1})}
In this study, we evaluate the performance of various backbones with and without the BeamVQ method. We choose 5 benchmarks (SWE(u), RBC, NSE, Prometheus, SEVIR) and use MSE as the evaluation metric. Table~\ref{tab:mainres} and Figure~\ref{fig:improve} summarize these results. Table~\ref{tab:mainres} shows the performance of different backbones on the five benchmarks. It is evident that all backbones achieve significantly lower MSE with the BeamVQ method. For example, the ResNet model's error in the SWE(u) benchmark decreases from 0.0076 to 0.0033, an improvement of 56.58\%. The ConvLSTM model's error in the RBC benchmark drops from 0.2726 to 0.0868, an improvement of 68.15\%. The Earthformer model's error in the NSE benchmark falls from 1.8720 to 0.1202, an improvement of 93.58\%. The overall improvement percentages for each benchmark are: SWE(u) 39.08\%, RBC 18.97\%, NSE 35.83\%, Prometheus 33.65\%, and SEVIR 35.27\%. These significant improvements demonstrate the effectiveness of the BeamVQ method in enhancing the physical consistency and prediction accuracy of spatiotemporal forecasting backbones. Figure~\ref{fig:improve} further illustrates the improvement percentages of different backbones across the benchmarks. 
It shows that the BeamVQ method significantly enhances model performance across all benchmarks. 

\begin{figure*}[t]
    \centering
    \includegraphics[width=\textwidth]{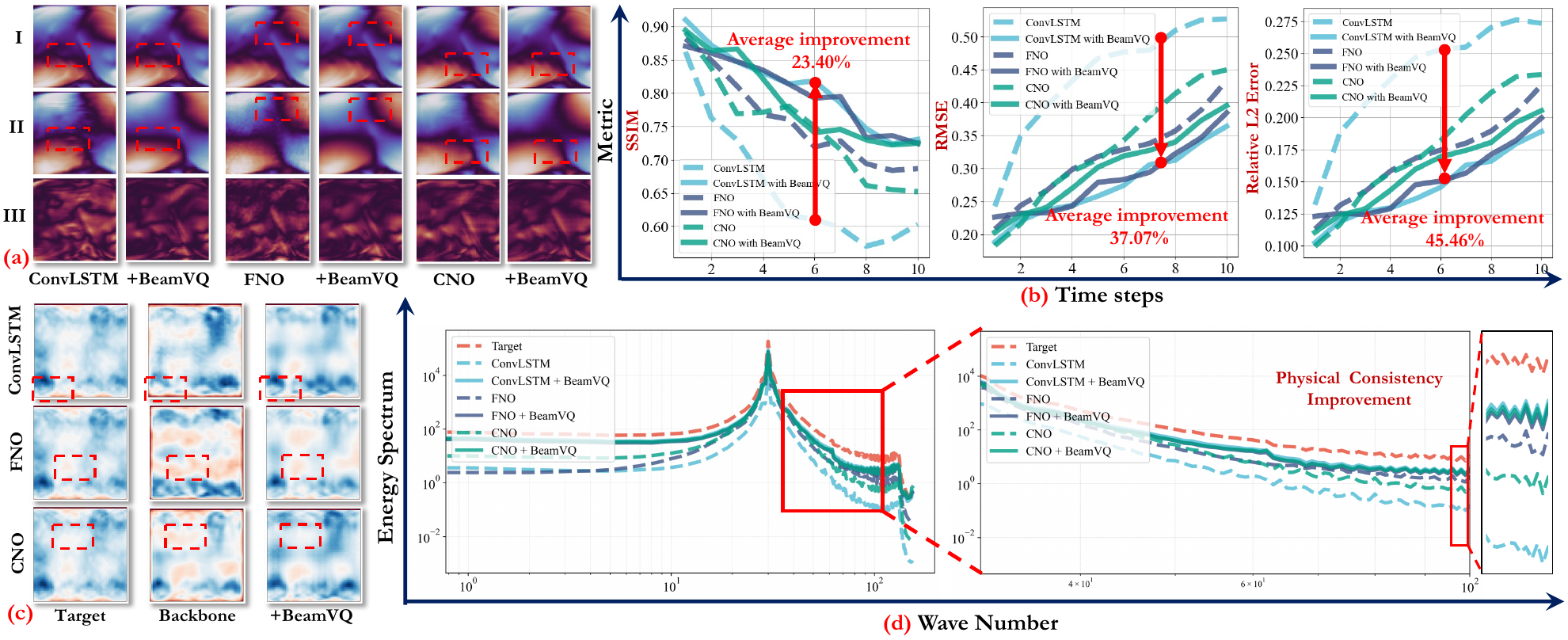}
    \caption{\textbf{The BeamVQ plugin improves physical consistency and prediction accuracy.} \textcolor{red}{(a)} shows a visual comparison of the actual target, predicted results, and errors at different time steps. \textcolor{red}{(b)} displays the changes in SSIM, RMSE, and relative L2 error over time steps. \textcolor{red}{(c)} compares the turbulent TKE. \textcolor{red}{(d)} presents the energy spectrum at different wavenumbers.}
    \label{fig:phy}
    \vspace{-10pt}
\end{figure*}

\begin{figure*}[t]
    \centering
    \includegraphics[width=\textwidth]{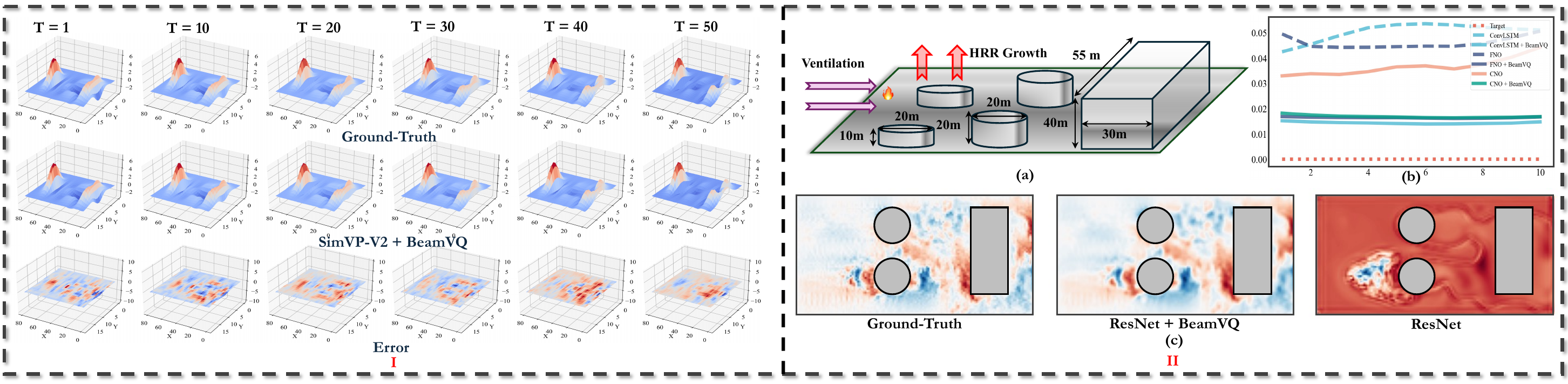}
    \caption{
    \textcolor{red}{I.} 3D visualization of the SWE(h), showing Ground-truth, SimVP-V2+BeamVQ predictions, and Error at T=1, 10, 20, 30, 40, 50. The first row shows Ground-truth, the second SimVP-V2+BeamVQ predictions, and the third Error. \textcolor{red}{II.} A case study. Building fire simulation with ventilation settings added to Wu's Prometheus~\cite{wu2024prometheus}. (a) Layout and HRR growth. (b) Comparison of physical metrics for different methods. (c) Ground-truth, ResNet+BeamVQ, and ResNet predictions.
    }
    \label{fig:case} 
    \vspace{-20pt}
\end{figure*}

\subsection{\method{} Boosts Physical Alignment (\textit{RQ2})}
\begin{wraptable}{r}{0.48\textwidth}
\vspace{-10pt}
  \centering
  \begin{sc}
  \caption{We compare different backbones on the SWE Benchmark for Long-term Forecasting.}
  \vspace{-5pt}
  \label{tab:time}
  \resizebox{\linewidth}{!}{%
    \begin{tabular}{cc|c|c|c}
    \toprule
    \multicolumn{1}{c}{Model} & & \multicolumn{1}{c|}{SWE (u)} & \multicolumn{1}{c|}{SWE (v)} & \multicolumn{1}{c}{SWE (h)} \\ 
    \midrule
    \multicolumn{1}{c}{\multirow{2}{*}{Simvp-v2}} & Ori  & 0.0187 &0.0387  &0.0443    \\
    \multicolumn{1}{c}{} & +\method{}  & \textbf{0.0154} & \textbf{0.0342} & \textbf{0.0397} \\ 
    \midrule
    \multicolumn{1}{c}{\multirow{2}{*}{ConvLSTM}} & Ori & 0.0487 &0.0673  &0.0762    \\
    \multicolumn{1}{c}{} & +\method{} & \textbf{0.0321} &\textbf{0.0351}  &\textbf{0.0432}   \\ 
    \midrule
    \multicolumn{1}{c}{\multirow{2}{*}{FNO}} & Ori  & 0.0571 &0.0832  &0.0981 \\
    \multicolumn{1}{c}{} & +\method{}  & \textbf{0.0502} &\textbf{0.0653}  &\textbf{0.0911}  \\ 
    \midrule
    \multicolumn{1}{c}{\multirow{2}{*}{CNO}} & Ori  &  0.1283 &0.1422  &0.1987  \\
    \multicolumn{1}{c}{} & +\method{}  & \textbf{0.0621} &\textbf{0.0674}  & \textbf{0.0965} \\ 
    \midrule
    \end{tabular}%
  }
  \end{sc}
  \vspace{-15pt}
\end{wraptable}
Figure~\ref{fig:phy} shows \method{} significantly improves physical consistency and prediction accuracy. In Figure~\ref{fig:phy}(a), the visualization of the actual target, predicted results, and error at different time steps shows that the method with BeamVQ performs better in details and physical consistency, with smaller errors. Figure~\ref{fig:phy}(b) presents performance metrics of SSIM, RMSE, and relative L2 error. The BeamVQ method improves these metrics by 23.40\%, reduces them by 37.07\%, and 45.46\%, respectively, indicating stronger robustness in spatiotemporal dynamical system prediction. Figure~\ref{fig:phy}(c) compares turbulent kinetic energy (TKE). The BeamVQ method captures changes in TKE more accurately, especially in details and small-scale turbulent structures. Figure~\ref{fig:phy}(d) shows the energy spectrum at different wavenumbers. The BeamVQ method demonstrates better physical consistency in the high-wavenumber region, meaning more accurate predictions of small-scale vortices. Overall, the BeamVQ framework not only enhances numerical accuracy in predictions but also captures the essence of physical phenomena better, showing significant advantages.

\vspace{-5pt}
\subsection{\method{} Excels In Long-term Dynamic System Forecasting (\textit{RQ3})}   \vspace{-5pt}
\label{sec:long}

In the long-term forecasting experiments, we compare the performance of different backbone models on the SWE benchmark, evaluating the relative L2 error for three variables (U, V, and H). Our setup inputs 5 frames and predicts 50 frames. For the SimVP-v2 model, using \method{} reduces the relative L2 error for SWE (u) from 0.0187 to 0.0154, SWE (v) from 0.0387 to 0.0342, and SWE (h) from 0.0443 to 0.0397. We visualize SWE (h) in 3D as shown in Figure~\ref{fig:case} [\textcolor{red}{I}]. For the ConvLSTM model, applying \method{} reduces the relative L2 error for SWE (u) from 0.0487 to 0.0321, SWE (v) from 0.0673 to 0.0351, and SWE (h) from 0.0762 to 0.0432. For the FNO model, using \method{} reduces the relative L2 error for SWE (u) from 0.0571 to 0.0502, SWE (v) from 0.0832 to 0.0653, and SWE (h) from 0.0981 to 0.0911. Overall, \method{} significantly improves the long-term forecasting accuracy of different backbone models.

\subsection{Interpretation Analysis \& Ablation Study}
\begin{figure*}[h]
    \vspace{-10pt}
    \centering
    \includegraphics[width=\textwidth]{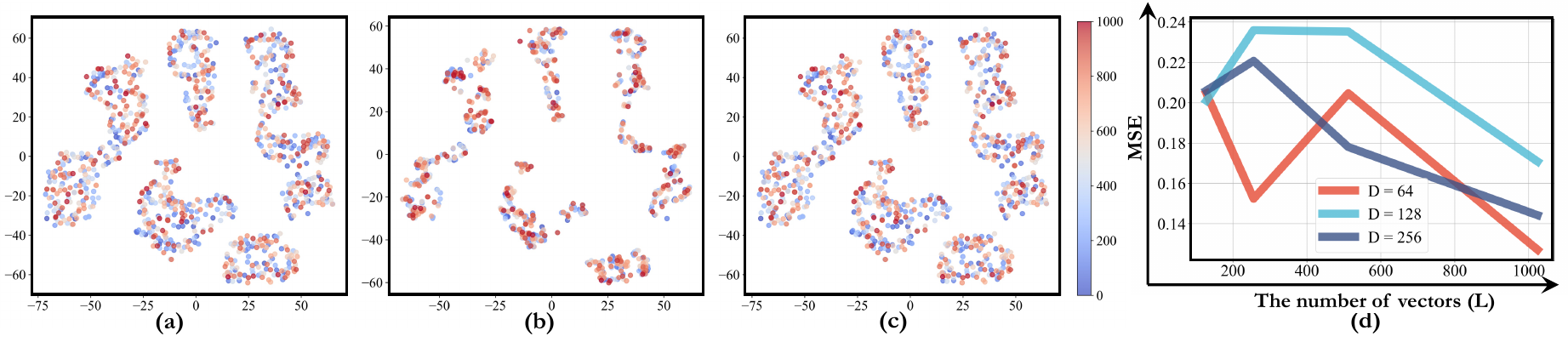}
    \vspace{-15pt}
    \caption{The t-SNE visualization in (a), (b), and (c) shows the Ground-truth, ConvLSTM and ConvLSTM+BeamVQ predictions, respectively. (d) shows the analysis of the Codebank parameters.}
    \label{fig:tsne} 
    \vspace{-15pt}
\end{figure*}

\textbf{Qualitative analysis using t-SNE.} Figure~\ref{fig:tsne} shows the t-SNE visualizations of the RBC dataset, including (a) ground truth labels, (b) ConvLSTM predictions, and (c) ConvLSTM+BeamVQ predictions. In (a), the ground truth labels display clear clusters, serving as a benchmark for evaluating model predictions. In (b), the ConvLSTM predictions show a more blurred clustering pattern with significant overlap, indicating that ConvLSTM struggles to capture the data structure accurately. In (c), the ConvLSTM+BeamVQ predictions exhibit clearer clusters that are closer to the ground truth, demonstrating that BeamVQ significantly enhances the ConvLSTM's predictive capability. Overall, the t-SNE results indicate that BeamVQ effectively improves the prediction accuracy and physical consistency of ConvLSTM on the RBC dataset, further validating BeamVQ's effectiveness.

\textbf{Analysis on Code bank.} We train FNO+BeamVQ on NSE for 100 epochs with a learning rate of 0.001 and a batch size of 100. The results are shown in the Figure~\ref{fig:tsne}. In the VQVAE codebank dimension experiment, we find that increasing the number of vectors $L$ significantly reduces MSE. Especially, when $L = 1024$ and $D = 64$, the MSE drops to the lowest value of 0.1271. Although MSE fluctuates more with $L = 256$ and $L = 512$, overall, increasing $L$ helps improve model accuracy. The training loss curves show that most combinations quickly decrease and stabilize within the first 20 epochs. Notably, the combination of $L = 512$ and $D = 128$ shows higher stability during training. In conclusion, the combination of $L = 1024$ and $D = 64$ is best for reducing MSE, while $L = 512$ and $D = 128$ perform well in training stability.

\begin{wraptable}{r}{0.5\textwidth}
\vspace{-5pt}
  \centering
  \begin{sc}
  \caption{Ablation studies on the NSE benchmark.}  
  \label{tab:ablation}
  \resizebox{\linewidth}{!}{%
    \begin{tabular}{l|c|c}
    \toprule
    Variants &MSE& TKE \\ 
    \midrule
    FNO  &0.2237  &0.3964     \\
    FNO+\method{} & \textbf{0.1005} & \textbf{0.1572}  \\ 
    FNO+\method{} (w/o BeamS)& 0.1207 & 0.2003  \\ 
    FNO+\method{} (w/o SelfT)  & 0.1118 & 0.1872\\ 
    FNO+\method{} (w. MSE)  & 0.1654 & 0.2847 \\  
    FNO+VQVAE  & 0.1872 & 0.3652 \\ 
    FNO+PINO  & 0.1249 & 0.2342  \\  
    \midrule
    \end{tabular}%
  }
  \end{sc}
  \vspace{-10pt}
\end{wraptable}
\textbf{Ablation Study.} We use NSE as a benchmark and conduct ablation experiments with FNO. The variants are: (I) FNO: The base model. (II) FNO+BeamVQ: Adds the BeamVQ module to FNO. (III) FNO+BeamVQ (w/o Beamsearch): Adds the BeamVQ module to FNO without Beamsearch. (IV) FNO+BeamVQ (w/o self-Training): Adds the BeamVQ module to FNO without self-training. (V) FNO+\method{} (w MSE): Uses MSE metrics instead of physical metrics for filtering. (VI) FNO+VQVAE: Uses VQVAE for discretization on top of FNO. (VII) FNO+PINO~\cite{10.1145/3648506}: Adds the Physics-Informed Neural Operator (PINO) to FNO, integrating physical constraints. The results in Table~\ref{tab:ablation} show that the base model FNO has an MSE of 0.2237 and a TKE relative error of 0.3964. Adding the BeamVQ module reduces the MSE to 0.1005 and the TKE relative error to 0.1572. Removing the Beamsearch strategy and self-training decreases performance but still improves over the base model, with MSEs of 0.1207 and 0.1118 and TKE relative errors of 0.2003 and 0.1872, respectively. Using VQVAE and PINO for optimization results in MSEs of 0.1872 and 0.1249 and TKE relative errors of 0.3652 and 0.2342. These results indicate that the BeamVQ module significantly improves the model's physical consistency and prediction accuracy.

\section{Conclusion and Future Work}
We propose BeamVQ, a novel probabilistic model to align space-time forecasting with Physics-aware metrics through self-training. 
BeamVQ uses a code bank to discretize continuous state space, converting any encoder-decoder model into a probabilistic model for beam search. 
It filters high-quality sequences based on Physics-aware scores to expand the dataset for self-training. 
As a general method, \method{} can work with different backbone networks, evaluation metrics, and Physical applications.
Our experiments show that BeamVQ consistently improves prediction quality and physical consistency, even with weak backbone models. 
In the future,
we can explore to 
combine with Physical regularization loss,
consider more large and complex datasets
and use more non-differentiable and differentiable metrics.

\clearpage

\bibliographystyle{plain}
\bibliography{references.bib}

\begin{thebibliography}{10}

\bibitem{bi2023accurate}
Kaifeng Bi, Lingxi Xie, Hengheng Zhang, Xin Chen, Xiaotao Gu, and Qi~Tian.
\newblock Accurate medium-range global weather forecasting with 3d neural networks.
\newblock {\em Nature}, 619(7970):533--538, 2023.

\bibitem{boulanger2013high}
Nicolas Boulanger-Lewandowski, Yoshua Bengio, and Pascal Vincent.
\newblock High-dimensional sequence transduction.
\newblock In {\em 2013 IEEE international conference on acoustics, speech and signal processing}, pages 3178--3182. IEEE, 2013.

\bibitem{cranmer2020lagrangian}
Miles Cranmer, Sam Greydanus, Stephan Hoyer, Peter Battaglia, David Spergel, and Shirley Ho.
\newblock Lagrangian neural networks.
\newblock {\em arXiv preprint arXiv:2003.04630}, 2020.

\bibitem{dong2023raft}
Hanze Dong, Wei Xiong, Deepanshu Goyal, Yihan Zhang, Winnie Chow, Rui Pan, Shizhe Diao, Jipeng Zhang, Kashun Shum, and Tong Zhang.
\newblock Raft: Reward ranked finetuning for generative foundation model alignment.
\newblock {\em arXiv preprint arXiv:2304.06767}, 2023.

\bibitem{dosovitskiy2021an}
Alexey Dosovitskiy, Lucas Beyer, Alexander Kolesnikov, Dirk Weissenborn, Xiaohua Zhai, Thomas Unterthiner, Mostafa Dehghani, Matthias Minderer, Georg Heigold, Sylvain Gelly, Jakob Uszkoreit, and Neil Houlsby.
\newblock An image is worth 16x16 words: Transformers for image recognition at scale.
\newblock In {\em International Conference on Learning Representations}, 2021.

\bibitem{gao2024prediff}
Zhihan Gao, Xingjian Shi, Boran Han, Hao Wang, Xiaoyong Jin, Danielle Maddix, Yi~Zhu, Mu~Li, and Yuyang~Bernie Wang.
\newblock Prediff: Precipitation nowcasting with latent diffusion models.
\newblock {\em Advances in Neural Information Processing Systems}, 36, 2023.

\bibitem{gao2022earthformer}
Zhihan Gao, Xingjian Shi, Hao Wang, Yi~Zhu, Yuyang~Bernie Wang, Mu~Li, and Dit-Yan Yeung.
\newblock Earthformer: Exploring space-time transformers for earth system forecasting.
\newblock {\em Advances in Neural Information Processing Systems}, 35:25390--25403, 2022.

\bibitem{greydanus2019hamiltonian}
Samuel Greydanus, Misko Dzamba, and Jason Yosinski.
\newblock Hamiltonian neural networks.
\newblock {\em Advances in neural information processing systems}, 32, 2019.

\bibitem{griebel1998numerical}
Michael Griebel, Thomas Dornseifer, and Tilman Neunhoeffer.
\newblock {\em Numerical simulation in fluid dynamics: a practical introduction}.
\newblock SIAM, 1998.

\bibitem{guen2020disentangling}
Vincent~Le Guen and Nicolas Thome.
\newblock Disentangling physical dynamics from unknown factors for unsupervised video prediction.
\newblock In {\em Proceedings of the IEEE/CVF conference on computer vision and pattern recognition}, pages 11474--11484, 2020.

\bibitem{guo2023beam}
Jeff Guo and Philippe Schwaller.
\newblock Beam enumeration: Probabilistic explainability for sample efficient self-conditioned molecular design.
\newblock {\em ICLR}, 2024.

\bibitem{gutzwiller1970energy}
Martin~C Gutzwiller.
\newblock Energy spectrum according to classical mechanics.
\newblock {\em Journal of Mathematical Physics}, 11(6):1791--1806, 1970.

\bibitem{hansen2023learning}
Derek Hansen, Danielle~C Maddix, Shima Alizadeh, Gaurav Gupta, and Michael~W Mahoney.
\newblock Learning physical models that can respect conservation laws.
\newblock In {\em International Conference on Machine Learning}, pages 12469--12510. PMLR, 2023.

\bibitem{he2016deep}
Kaiming He, Xiangyu Zhang, Shaoqing Ren, and Jian Sun.
\newblock Deep residual learning for image recognition.
\newblock In {\em Proceedings of the IEEE conference on computer vision and pattern recognition}, pages 770--778, 2016.

\bibitem{jouvet2009numerical}
Guillaume Jouvet, Matthias Huss, Heinz Blatter, Marco Picasso, and Jacques Rappaz.
\newblock Numerical simulation of rhonegletscher from 1874 to 2100.
\newblock {\em Journal of Computational Physics}, 228(17):6426--6439, 2009.

\bibitem{karnopp2012system}
Dean~C Karnopp, Donald~L Margolis, and Ronald~C Rosenberg.
\newblock {\em System dynamics: modeling, simulation, and control of mechatronic systems}.
\newblock John Wiley \& Sons, 2012.

\bibitem{karpatne2017theory}
Anuj Karpatne, Gowtham Atluri, James~H Faghmous, Michael Steinbach, Arindam Banerjee, Auroop Ganguly, Shashi Shekhar, Nagiza Samatova, and Vipin Kumar.
\newblock Theory-guided data science: A new paradigm for scientific discovery from data.
\newblock {\em IEEE Transactions on knowledge and data engineering}, 29(10):2318--2331, 2017.

\bibitem{kingma2014adam}
Diederik~P Kingma and Jimmy Ba.
\newblock Adam: A method for stochastic optimization.
\newblock {\em arXiv preprint arXiv:1412.6980}, 2014.

\bibitem{krishnapriyan2021characterizing}
Aditi Krishnapriyan, Amir Gholami, Shandian Zhe, Robert Kirby, and Michael~W Mahoney.
\newblock Characterizing possible failure modes in physics-informed neural networks.
\newblock {\em Advances in Neural Information Processing Systems}, 34:26548--26560, 2021.

\bibitem{lam2022graphcast}
Remi Lam, Alvaro Sanchez-Gonzalez, Matthew Willson, Peter Wirnsberger, Meire Fortunato, Ferran Alet, Suman Ravuri, Timo Ewalds, Zach Eaton-Rosen, Weihua Hu, et~al.
\newblock Graphcast: Learning skillful medium-range global weather forecasting.
\newblock {\em arXiv preprint arXiv:2212.12794}, 2022.

\bibitem{li2020fourier}
Zongyi Li, Nikola Kovachki, Kamyar Azizzadenesheli, Burigede Liu, Kaushik Bhattacharya, Andrew Stuart, and Anima Anandkumar.
\newblock Fourier neural operator for parametric partial differential equations.
\newblock {\em arXiv preprint arXiv:2010.08895}, 2020.

\bibitem{li2021physics}
Zongyi Li, Hongkai Zheng, Nikola Kovachki, David Jin, Haoxuan Chen, Burigede Liu, Kamyar Azizzadenesheli, and Anima Anandkumar.
\newblock Physics-informed neural operator for learning partial differential equations.
\newblock {\em ACM/JMS Journal of Data Science}, 2021.

\bibitem{10.1145/3648506}
Zongyi Li, Hongkai Zheng, Nikola Kovachki, David Jin, Haoxuan Chen, Burigede Liu, Kamyar Azizzadenesheli, and Anima Anandkumar.
\newblock Physics-informed neural operator for learning partial differential equations.
\newblock {\em ACM / IMS J. Data Sci.}, feb 2024.
\newblock Just Accepted.

\bibitem{long2018pde}
Zichao Long, Yiping Lu, Xianzhong Ma, and Bin Dong.
\newblock {PDE}-net: Learning {PDE}s from data.
\newblock In {\em International conference on machine learning}, pages 3208--3216. PMLR, 2018.

\bibitem{nagata2013turbulence}
Kouji Nagata, Yasuhiko Sakai, Takuto Inaba, Hiroki Suzuki, Osamu Terashima, and Hiroyuki Suzuki.
\newblock Turbulence structure and turbulence kinetic energy transport in multiscale/fractal-generated turbulence.
\newblock {\em Physics of Fluids}, 25(6), 2013.

\bibitem{orszag1974numerical}
Steven~A Orszag and Moshe Israeli.
\newblock Numerical simulation of viscous incompressible flows.
\newblock {\em Annual Review of Fluid Mechanics}, 6(1):281--318, 1974.

\bibitem{paszke2019pytorch}
Adam Paszke, Sam Gross, Francisco Massa, Adam Lerer, James Bradbury, Gregory Chanan, Trevor Killeen, Zeming Lin, Natalia Gimelshein, Luca Antiga, et~al.
\newblock Pytorch: An imperative style, high-performance deep learning library.
\newblock {\em Advances in neural information processing systems}, 32, 2019.

\bibitem{pathak2022fourcastnet}
Jaideep Pathak, Shashank Subramanian, Peter Harrington, Sanjeev Raja, Ashesh Chattopadhyay, Morteza Mardani, Thorsten Kurth, David Hall, Zongyi Li, Kamyar Azizzadenesheli, et~al.
\newblock Fourcastnet: A global data-driven high-resolution weather model using adaptive fourier neural operators.
\newblock {\em arXiv preprint arXiv:2202.11214}, 2022.

\bibitem{pukrushpan2004control}
Jay~T Pukrushpan, Anna~G Stefanopoulou, and Huei Peng.
\newblock {\em Control of fuel cell power systems: principles, modeling, analysis and feedback design}.
\newblock Springer Science \& Business Media, 2004.

\bibitem{raissi2019physics}
Maziar Raissi, Paris Perdikaris, and George~E Karniadakis.
\newblock Physics-informed neural networks: A deep learning framework for solving forward and inverse problems involving nonlinear partial differential equations.
\newblock {\em Journal of Computational physics}, 378:686--707, 2019.

\bibitem{raonic2024convolutional}
Bogdan Raonic, Roberto Molinaro, Tim De~Ryck, Tobias Rohner, Francesca Bartolucci, Rima Alaifari, Siddhartha Mishra, and Emmanuel de~B{\'e}zenac.
\newblock Convolutional neural operators for robust and accurate learning of pdes.
\newblock {\em Advances in Neural Information Processing Systems}, 36, 2023.

\bibitem{razavi2019generating}
Ali Razavi, Aaron Van~den Oord, and Oriol Vinyals.
\newblock Generating diverse high-fidelity images with vq-vae-2.
\newblock {\em Advances in neural information processing systems}, 32, 2019.

\bibitem{rogallo1984numerical}
Robert~S Rogallo and Parviz Moin.
\newblock Numerical simulation of turbulent flows.
\newblock {\em Annual review of fluid mechanics}, 16(1):99--137, 1984.

\bibitem{shi2015convolutional}
Xingjian Shi, Zhourong Chen, Hao Wang, Dit-Yan Yeung, Wai-Kin Wong, and Wang-chun Woo.
\newblock Convolutional lstm network: A machine learning approach for precipitation nowcasting.
\newblock {\em Advances in neural information processing systems}, 28, 2015.

\bibitem{steinbiss1994improvements}
Volker Steinbiss, Bach-Hiep Tran, and Hermann Ney.
\newblock Improvements in beam search.
\newblock In {\em ICSLP}, volume~94, pages 2143--2146, 1994.

\bibitem{takamoto2022pdebench}
Makoto Takamoto, Timothy Praditia, Raphael Leiteritz, Daniel MacKinlay, Francesco Alesiani, Dirk Pfl{\"u}ger, and Mathias Niepert.
\newblock Pdebench: An extensive benchmark for scientific machine learning.
\newblock {\em Advances in Neural Information Processing Systems}, 35:1596--1611, 2022.

\bibitem{tan2022simvp}
Cheng Tan, Zhangyang Gao, Siyuan Li, and Stan~Z Li.
\newblock Simvp: Towards simple yet powerful spatiotemporal predictive learning.
\newblock {\em arXiv preprint arXiv:2211.12509}, 2022.

\bibitem{tan2023temporal}
Cheng Tan, Zhangyang Gao, Lirong Wu, Yongjie Xu, Jun Xia, Siyuan Li, and Stan~Z Li.
\newblock Temporal attention unit: Towards efficient spatiotemporal predictive learning.
\newblock In {\em Proceedings of the IEEE/CVF Conference on Computer Vision and Pattern Recognition}, pages 18770--18782, 2023.

\bibitem{tuckerman1989divergence}
Laurette~S Tuckerman.
\newblock Divergence-free velocity fields in nonperiodic geometries.
\newblock {\em Journal of Computational Physics}, 80(2):403--441, 1989.

\bibitem{van2017neural}
Aaron Van Den~Oord, Oriol Vinyals, et~al.
\newblock Neural discrete representation learning.
\newblock {\em Advances in neural information processing systems}, 30, 2017.

\bibitem{veillette2020sevir}
Mark Veillette, Siddharth Samsi, and Chris Mattioli.
\newblock Sevir: A storm event imagery dataset for deep learning applications in radar and satellite meteorology.
\newblock {\em Advances in Neural Information Processing Systems}, 33:22009--22019, 2020.

\bibitem{wang20222}
Chuwei Wang, Shanda Li, Di~He, and Liwei Wang.
\newblock Is l2 physics informed loss always suitable for training physics informed neural network?
\newblock {\em Advances in Neural Information Processing Systems}, 35:8278--8290, 2022.

\bibitem{wang2020towards}
Rui Wang, Karthik Kashinath, Mustafa Mustafa, Adrian Albert, and Rose Yu.
\newblock Towards physics-informed deep learning for turbulent flow prediction.
\newblock In {\em Proceedings of the 26th ACM SIGKDD international conference on knowledge discovery \& data mining}, pages 1457--1466, 2020.

\bibitem{wang2018eidetic}
Yunbo Wang, Lu~Jiang, Ming-Hsuan Yang, Li-Jia Li, Mingsheng Long, and Li~Fei-Fei.
\newblock Eidetic 3d lstm: A model for video prediction and beyond.
\newblock In {\em International conference on learning representations}, 2018.

\bibitem{wang2022predrnn}
Yunbo Wang, Haixu Wu, Jianjin Zhang, Zhifeng Gao, Jianmin Wang, S~Yu Philip, and Mingsheng Long.
\newblock Predrnn: A recurrent neural network for spatiotemporal predictive learning.
\newblock {\em IEEE Transactions on Pattern Analysis and Machine Intelligence}, 45(2):2208--2225, 2022.

\bibitem{wu2024prometheus}
Hao Wu, Huiyuan Wang, Kun Wang, Weiyan Wang, Changan Ye, Yangyu Tao, Chong Chen, Xian-Sheng Hua, and Xiao Luo.
\newblock Prometheus: Out-of-distribution fluid dynamics modeling with disentangled graph ode.
\newblock In {\em Proceedings of the 41st International Conference on Machine Learning}, page PMLR 235, Vienna, Austria, 2024. PMLR.

\bibitem{wu2024earthfarsser}
Hao Wu, Shilong Wang, Yuxuan Liang, Zhengyang Zhou, Wei Huang, Wei Xiong, and Kun Wang.
\newblock Earthfarseer: Versatile spatio-temporal dynamical systems modeling in one model.
\newblock {\em arXiv e-prints}, pages arXiv--2312, 2023.

\bibitem{wu2023pastnet}
Hao Wu, Wei Xion, Fan Xu, Xiao Luo, Chong Chen, Xian-Sheng Hua, and Haixin Wang.
\newblock Pastnet: Introducing physical inductive biases for spatio-temporal video prediction.
\newblock {\em arXiv preprint arXiv:2305.11421}, 2023.

\bibitem{wu2024neural}
Hao Wu, Shuyi Zhou, Xiaomeng Huang, and Wei Xiong.
\newblock Neural manifold operators for learning the evolution of physical dynamics, 2024.

\bibitem{yuan2024self}
Weizhe Yuan, Richard~Yuanzhe Pang, Kyunghyun Cho, Sainbayar Sukhbaatar, Jing Xu, and Jason Weston.
\newblock Self-rewarding language models.
\newblock {\em arXiv preprint arXiv:2401.10020}, 2024.

\bibitem{zhang2023skilful}
Yuchen Zhang, Mingsheng Long, Kaiyuan Chen, Lanxiang Xing, Ronghua Jin, Michael~I Jordan, and Jianmin Wang.
\newblock Skilful nowcasting of extreme precipitation with nowcastnet.
\newblock {\em Nature}, 619(7970):526--532, 2023.

\end{thebibliography}


\clearpage
\appendix

\section{The Proposed \method{} Algorithm}

\textbf{Self-Training with Increasing Frequency:} This algorithm gradually increases the frequency of self-training. In the initial phase ($t < \mathcal{E}_1$), it updates the model using only the original training data. In the middle phase ($\mathcal{E}_1 \leq t < \mathcal{E}_2$), it incorporates high-quality self-training data every 50 epochs. In the final phase ($t \geq \mathcal{E}_2$), it adds high-quality self-training data every 10 epochs to enhance the model's physical consistency. As shown in algorithm~\ref{alg:1}.

\textbf{Beam Search by Vector Quantization (\method{}):} This algorithm optimizes space-time forecasting models using beam search with vector quantization. It samples batches from the original training data, encodes each input, and uses beam search to generate multiple candidate states. It evaluates these candidates based on physics-aware scores, retains the top K states for further search, selects the highest-scoring sequence as the output, and adds high-quality outputs to the self-training dataset, continuously updating the model. As shown in algorithm~\ref{alg:2}.

\begin{algorithm}[H]
    \caption{Self-Training with Increasing Frequency}
    \begin{algorithmic}[1]
    \REQUIRE Training dataset $\mathcal{D} = (\mathcal{X}, \mathcal{Y})$, high-quality self-training dataset $\mathcal{D}_{\text{high}} = (\mathcal{X}_{\text{high}}, \mathcal{Y}_{\text{high}})$, total epochs $T = 500$, learning rate $\eta$, epochs $\mathcal{E}_1 = 100$ and $\mathcal{E}_2 = 200$
    \STATE Initialize parameters $\theta^{(0)}$
    \FOR{$t = 0$ to $T-1$}
    \IF{$t < \mathcal{E}_1$}
        \STATE Update Eqn.~\ref{eq:ori}
    \ELSIF{$\mathcal{E}_1 \leq t < \mathcal{E}_2$}
        \STATE $f(t) \leftarrow 50$
        \STATE Update Eqn.~\ref{eq:self}
    \ELSE
        \STATE $f(t) \leftarrow 10$
        \STATE Update Eqn.~\ref{eq:self}
    \ENDIF
\ENDFOR
\end{algorithmic}\label{alg:1}
\end{algorithm}

\begin{algorithm}[h]
\caption{Beam Search by Vector Quantization (\method{})}
\begin{algorithmic}[1]
\REQUIRE Training dataset $\mathcal{D} = (\mathcal{X}, \mathcal{Y})$, total epochs $T$, learning rate $\eta$, beam size $K$, Physics-aware threshold $\text{threshold}$
\STATE Initialize parameters $\theta^{(0)}$
\STATE Initialize high-quality self-training dataset $\mathcal{D}_{\text{high}} = \emptyset$
\FOR{$t = 0$ to $T-1$}
    \STATE $\mathcal{D}_{\text{batch}} = \mathcal{D} \cup \mathcal{D}_{\text{high}}$
    \STATE Sample a mini-batch $(\mathcal{X}_{\text{batch}}, \mathcal{Y}_{\text{batch}}) \subset \mathcal{D}_{\text{batch}}$
    \FOR{each $(\mathcal{X}, \mathcal{Y})$ in $(\mathcal{X}_{\text{batch}}, \mathcal{Y}_{\text{batch}})$}
        \STATE $\mathcal{Z} = E_{\phi}(\mathcal{X})$
        \FOR{each step in beam search}
            \STATE Sample $K$ candidate latent states $\{e_{k_1}, \ldots, e_{k_K}\}$ based on top-K closest vectors
            \STATE Decode each candidate state $s_i$ to the output sequence $\mathcal{Y}_i = D_{\varphi}(s_i), \forall i \in [K]$
            \STATE Evaluate Physics-aware scores $F(\mathcal{Y}_i)$
            \STATE Retain top-K sequences with highest Physics-aware scores for next step
        \ENDFOR
        \STATE Select the highest-scored sequence $\mathcal{Y}^*$
        \IF{$F(\mathcal{Y}^*) \geq \text{threshold}$}
            \STATE Add $(\mathcal{X}, \mathcal{Y}^*)$ to $\mathcal{D}_{\text{high}}$
        \ENDIF
    \ENDFOR
    \STATE Compute loss $\mathcal{L}$ on $(\mathcal{X}_{\text{batch}}, \mathcal{Y}_{\text{batch}})$ and update parameters:
    \STATE $\theta^{(t+1)} = \theta^{(t)} - \eta \nabla_\theta \mathcal{L}(\mathcal{X}_{\text{batch}}, \mathcal{Y}_{\text{batch}}; \theta^{(t)})$
\ENDFOR

\end{algorithmic}\label{alg:2}
\end{algorithm}

\end{document}